\documentclass[11pt]{article}
\usepackage{fullpage}
\usepackage{amsmath, amsthm, slashed}
\usepackage[left=1in,top=1in,right=1in,bottom=1in,headheight=3ex,headsep=3ex]{geometry}
\usepackage{graphicx}
\usepackage{float}

\usepackage{indentfirst}

\usepackage{amsmath,amssymb,amsfonts,amsthm,amsbsy,amstext}
\usepackage[authordate,backend=b
iber]{biblatex-chicago}
\usepackage{graphicx}
\usepackage{wrapfig}
\usepackage{multicol}
\usepackage{geometry}
\usepackage{amssymb}
\usepackage{mathtools}
\usepackage{siunitx}
\usepackage{color,soul}

\usepackage{times}
\usepackage[T1]{fontenc}
\usepackage[mmddyyyy]{datetime}% http://ctan.org/pkg/datetime

\usepackage{setspace}
\usepackage{multicol}
\usepackage{url}
\usepackage{lastpage}
\usepackage{amsmath}
\usepackage{layout}
\usepackage{hyperref}
\usepackage{cleveref}
\usepackage{epigraph} 
\usepackage{algorithm}
\usepackage{algpseudocode}
\usepackage{array, xcolor}
\usepackage{color}
\usepackage{subfig}
\definecolor{clemsonorange}{HTML}{EA6A20}
\hypersetup{colorlinks,breaklinks,linkcolor=blue,urlcolor=blue,anchorcolor=clemsonorange,citecolor=blue}

\usepackage{url}

\newcommand{\R}{\mathbb R}

\newcommand{\E}{\mathbb E}

\newcommand{\comment}[1]{}

\theoremstyle{definition}

\crefname{assumption}{assumption}{assumptions}
\Crefname{assumption}{Assumption}{Assumptions}
\crefname{assumptionalt}{assumption}{assumptions}
\Crefname{assumptionalt}{Assumption}{Assumptions}
\usepackage{tkz-graph} 
\usetikzlibrary{arrows,automata, positioning}
\usetikzlibrary{shapes.geometric}%   
\usepackage{titlesec}
\titleformat*{\section}{\large\bfseries}
\titleformat*{\subsection}{\bfseries}
\titleformat*{\subsubsection}{\bfseries}
\titleformat*{\paragraph}{\bfseries}
\titleformat*{\subparagraph}{\bfseries}

\title{Using LLMs to Directly Guess Conditional Expectations Can Improve Efficiency in Causal Estimation}
\author{Chris Engh and P. M. Aronow \\ Yale University}

\begin{document}

\maketitle
\begin{abstract}
    We propose a simple yet effective use of LLM-powered AI tools to improve causal estimation. In double machine learning, the accuracy of causal estimates of the effect of a treatment on an outcome in the presence of a high-dimensional confounder depends on the performance of estimators of conditional expectation functions. We show that predictions made by generative models trained on historical data can be used to improve the performance of these estimators relative to approaches that solely rely on adjusting for embeddings extracted from these models. We argue that the historical knowledge and reasoning capacities associated with these generative models can help overcome curse-of-dimensionality problems in causal inference problems. We consider a case study using a small dataset of online jewelry auctions, and demonstrate that inclusion of LLM-generated guesses as predictors can improve efficiency in estimation.
\end{abstract}

% A common approach for estimating the causal effect of a treatment $D$ on an outcome $Y$ in the presence of high-dimensional confounders $W$ is to control for a (possibly fine-tuned) embedding of $W$ generated from an LLM. [[citation]]\\

\section{Introduction}

Researchers often face causal inference problems in which they wish to estimate the causal effect of a treatment $D$ on some outcome $Y$, adjusting for confounders $W$. However, when $W$ is high-dimensional and potentially unstructured and multimodal (e.g., text, image data), doing so can pose serious statistical difficulties. Even when $W$ is sufficient to control for confounding, the researcher is still faced with a difficult curse-of-dimensionality problem. Without additional information with which to reduce the complexity of the problem, it is generally impossible to attain a causal estimator with good statistical guarantees \parencite{robins1997toward}. \\

One common approach is to use an LLM to extract embeddings from $W$, and attempt to control for the embeddings as part of a covariate control strategy. Even though this reduces the complexity of the problem, estimation remains statistically difficult: the embedding typically is high-dimensional, and the functional form relating the embeddings to $Y$ or $D$ may be hard to characterize or estimate nonparametrically. In short, the problem is that there may not enough observations to support precise nonparametric supervised learning.\\

We suggest that modern LLM-based AI tools can be used to improve causal estimation by bringing in additional information derived from historical knowledge and reasoning. This entails simply asking the LLM to make predictions about the values of $Y$ and $D$ based on supplied data. This type of query may be able to yield good candidate guesses for the conditional expectation functions of $Y$ and $D$, due to the documented ability of LLMs to make statistical predictions based on historical data in this manner \parencite{gruver2023large}. And because modern reasoning models can refer to $W$ multiple times \parencite{openai_reasoning_guide}, their predictions can incorporate information about $W$ that would not be reflected in the embedding of the original data. \\

We demonstrate that these LLM-generated guesses can be used to improve the efficiency of causal estimation under the double machine learning (DML) framework \parencite{chernozhukov2018double}. The efficiency of DML critically depends on how well two conditional expectation functions ($\E[Y|W]$, $\E[D|W]$) are estimated. We propose a procedure that essentially entails directly incorporating the LLM-generated guesses as predictors alongside the usual embeddings in the estimation of these two nuisance functions. In an empirical application to online jewelry auctions, we show that this procedure does in fact improve efficiency in causal estimation relative to an embeddings-only approach. 

%The convergence rate of doubly-robust estimators depend on the convergence of the estimators of the nuisance functions $\E[D|W]$ and $\E[Y|W]$.\\

%To improve the performance of causal estimators, we suggest that 

% propose controlling for synthetic variables generated by querying large language models. For example, if one seeks to infer the causal effect of a treatment $D$ on a health outcome $Y$ conditional on a patient's medical history $W$, one could control for the patient's age, existing medical conditions (0-1), subjective measures of health and well-being (1-10), family history (0-1), \textit{etc.}

%The simplest synthetic variables to generate would be guesses of the nuisance functions themselves; even such a query can substantially improve efficiency.\\

%As an illustration, we estimate the nuisances necessary for estimation of the causal effect of seller feedback scores on eBay jewelry auctions. 

\subsection{Related work}

This paper contributes to a growing literature on incorporating AI into statistical inference. In both structural and causal estimation, researchers have used AI to impute unobserved or latent variables. \textcite{ludwig2025large} discuss at length the credibility of applying AI to econometric models. In recent work, \textcite{gui2025leveragingllmsimproveexperimental} directly query LLMs in a similar manner to predict potential outcomes so as to improve the design of randomized experiments. Similarly, in causal estimation settings with unobserved confounders, AI is increasingly being used to generate controls with the aim of closing backdoor paths; \textcite{battaglia2024inference} highlight the potential and pitfalls of this approach and provide ways to ensure the validity of the approach in a linear regression setting. In causal machine learning, machine learning methods are commonly used to estimate the nuisance functions that are plugged into doubly robust estimators of causal effects \parencite{robins1994estimation,chernozhukov2018double}. The efficiency of these estimators depend on the ability of the predictive models to be estimated well. To this end, some recent papers fine-tune language/image representations or tailor neural architectures to improve nuisance estimation \parencite{klaassen2024doublemldeep,bach2024adventures,veljanovski2024doublelingo}; this stands in contrast to earlier work on producing low-dimensional causal embeddings aimed at capturing the confounding elements of a text \parencite{roberts2020textmatching,veitch2020adapting}. The main contribution of our paper is to improve the efficiency of DML estimation of causal parameters by proposing a simple yet effective use of AI to improve the estimation of complicated nuisance functions. This is especially relevant in (but not limited to) settings where $n$ is not sufficiently large to fine-tune embeddings to pick up on the key non-linearities involved in the nuisance functions.

% \noindent 
% \textit{Example: the effect of a treatment on survival time.} Take $Y$ to be the survival time of a patient after diagnosis, $D$ to be whether or not the patient undergoes an established treatment, and $W$ to be the patient's health records, which could contain low-dimensional items, such as the patient's other medical conditions, age, existing medications, propensity to take prescriptions, \textit{etc.}, but also higher-dimensional data including clinical notes, lab results and imagining data.\\

%\noindent 
%\textit{Example: the effect of sex on criminal sentencing.} Take $Y$ to be the prison sentence that a convict receives, $D$ to be the sex of the offender, and $W$ to be all documents related to the trial, which could contain low-dimensional items, such as the defendant's other characteristics, the crime, aggravating circumstances, heinousness, emotional resonance of victim impact statements, \textit{etc.}\\

\section{Case study: effect of seller reputation on auction prices in online jewelry auctions}

eBay is a website on which both small and large sellers can auction off items, such as jewelry. The seller's \textit{feedback score} is the number of positive ratings less the number of negative ratings. Our interest lies in estimating the effect of the feedback score on auction prices, adjusting for all available information in the content of the listing. See \textcite{tadelis16} for a review of the literature of seller reputation on auction prices.

This setting may be favorable for causal identification because the data can be exhaustive or nearly exhaustive with respect to the information available to bidders, potentially implying exogeneity of errors. However, even if  identification is straightforward, causal estimation may still be difficult because the content in a listing is high-dimensional and multimodal.

\subsection{Model}

 Let $D$ denote the quantile of an item's seller's feedback score and let $Y$ denote be the log of its auction price. We are interested in the effect of $D\in \R$ on $Y\in \R$, confounded by $W\in \R^{\dim(W)}$, which represents the text and image data in the item listing. For the purposes of this paper, we will consider the partially linear model \parencite{robinson1988root}; $Y$ and $D$ are generated according to
\begin{align*}
    Y_i=D_i\theta_0+g(W_i)+U_i&& \E[U_i|W_i,D_i]=0\\ 
    D_i=m(W_i)+V_i && \E[V_i|W_i]=0
\end{align*}
The statistician observes a single cross-sectional dataset $\{Y_i,D_i,W_i\}_{i\in [n]}$. 

\subsection{Data} 

We retrieved a list of the 720 most recent completed auctions on eBay under the search term ``gold necklace'' within the ``Fine Necklaces \& Pendants'' category, applying additional filters for i) auctions ii) completed items (i.e. auctions that have ended) and iii) items whose metal was gold. The auctions were completed between 15 September and 22 September, 2025. We then queried eBay's trading API for relevant listing information. We obtained text-based listing information using the using the \texttt{GetItem} call with the specification \texttt{includeitemspecifics}, which includes the seller's feedback score. We then masked the seller information such that neither the embeddings nor the LLM had access to it. In a separate \texttt{GetItem} call, we obtained the links to the images in the listing's image carousel. We obtained bid data from the \texttt{GetAllBidders} call, including the number of bidders and bid values in USD.\\

From the listing data, we filtered down to a dataset of $n=333$ listings for which at least one bid was made (and consequently, the item was sold). For the purposes of this case study, we set aside concerns about (differential) sample selection bias \parencite{livingston2005valuable}, and focus on the distributional effect on prices given sales. The raw data was used to generate summary statistics (\cref{summary}) then passed through to generate embeddings and LLM predictions to produce a secondary dataset from which we obtained our results.

%\noindent 
%\textbf{Estimation.} We follow the  approach to estimation. We use residual-on-residual regression to estimate $\theta$ in the partially linear model:
%\begin{align}
 %   \hat r =Y-\hat \E[Y|W]\\ 
%    \hat v=D-\hat\E[D|W]\\
%    \hat\theta = \frac{\sum \hat v_i\hat r_i}{\sum\hat v_i^2}
%\end{align}
%where $\hat\E[Y|W]$ and $\hat\E[D|W]$ are out-of-fold predictions. Details on estimating the predictions $\hat\E[Y|W]$ and $\hat\E[D|W]$ will be provided in the Application.

\subsection{Embeddings} We obtained embeddings of listing images using Google's Vertex multimodal image embedding model \texttt{multimodalembedding@001} ($d=1408$). If a listing contained more than one image, we $L^2$-normalized the embeddings of each image and averaged over them to obtain a single 1408-dimensional vector. Because we are averaging, we limited ourselves to the first twelve images, although eBay allows upwards of twenty-four.\footnote{As shown in \cref{summary}, 75\% of listings had fewer than nine images. We also limited the LLM to twelve images. The LLM reasoning outputs suggest that the LLM used the image data to gauge high-level features, such as professionalism, studio lighting, and wear, as opposed to sharp signals. Furthermore, these signals were second order to the LLM's estimate of the intrinsic value. } We obtained embeddings of the text data using Google's Gemini model \texttt{gemini-embedding-001} ($d=768$). 

\subsection{Generative LLM predictions} We use \texttt{gpt-5-mini-2025-08-07}, a multimodal model that can accept image links and text as inputs. We queried the model using the system prompts in \cref{prompt_price,prompt_fs} along with the listing's image links and getitem text and asked it to predict $Y$ and $D$, given the text and images $W$. We queried separately for $Y$ and $D$; for predicting $D$, we masked seller information from the text input. We provide sample outputs in \cref{sampleoutputs}. \\ 

We investigated the possibility of data leakage. The \texttt{getitem} call does not include any information about the bids. We queried GPT through OpenAI's API; the payload did not allow for access to the internet beyond accessing the image links. Furthermore, inspection of the model's reasoning outputs suggests that only the raw images and the text provided were used.

\subsection{Methods}  Let $X_i=\phi(W_i)\in\mathbb{R}^p$ denote a (fixed) concatenation of text and image embeddings, and let $\tilde Y_i,\tilde D_i$ be LLM-generated predictions of $Y_i$ and $D_i$, respectively. We use residual-on-residual regression with cross-fitting (DML) \parencite{chernozhukov2018double}. Partition the sample into $K=5$ folds $I_1,\dots,I_K$ of (approximately) equal size, and write $I_k^c$ for the complement of fold $k$. \\

First we consider the outcome regression, $g'(\cdot) = \mathbb{E}[Y\mid W=\cdot]$. For each fold $k$, choose a penalty $\lambda_{Y,k}$ by leave-one-out cross-validation (LOOCV) within the training data $I_k^c$. Then fit the following partially penalized LASSO, leaving the coefficient on the LLM guess $\tilde Y$ unpenalized:
\begin{equation*}
(\hat b_{Y,k},\hat\alpha_k,\hat\beta_{Y,k})
\in\arg\min_{b,\alpha,\beta\in\mathbb{R}^{1+p}}
\left\{
\frac{1}{|I_k^c|}\sum_{i\in I_k^c}\!\big(Y_i - b - \alpha \tilde Y_i - X_i^\top \beta\big)^2
\;+\; \lambda_{Y,k}\,\lVert \beta\rVert_1
\right\}.
\end{equation*} 
In our base specification, we do not regularize the coefficient on the LLM guess because it is low-dimensional and believed to be strongly associated with the outcome.\footnote{In an unreported supplemental analysis, we allow for the LLM guesses to be penalized, and it does not meaningfully change the results.} We then compute out-of-fold predictions for $i\in I_k$:
\begin{equation*}
\widehat g^{(-k)}(W_i)\;=\;\hat b_{Y,k}+\hat\alpha_k \tilde Y_i + X_i^\top \hat\beta_{Y,k}.
\end{equation*}

For the treatment regression $m'(\cdot)=\mathbb{E}[D\mid W=\cdot]$, we analogously select $\lambda_{D,k}$ by LOOCV on $I_k^c$ and fit
\begin{equation*}
(\hat b_{D,k},\hat\gamma_k,\hat\beta_{D,k})
\in\arg\min_{b,\gamma,\beta\in\mathbb{R}^{1+p}}
\left\{
\frac{1}{|I_k^c|}\sum_{i\in I_k^c}\!\big(D_i - b - \gamma \tilde D_i - X_i^\top \beta\big)^2
\;+\; \lambda_{D,k}\,\lVert \beta\rVert_1
\right\},
\end{equation*}
then compute out-of-fold predictions for $i\in I_k$:
\begin{equation*}
\widehat m^{(-k)}(W_i)\;=\;\hat b_{D,k}+\hat\gamma_k \tilde D_i + X_i^\top \hat\beta_{D,k}.
\end{equation*}
We also estimate a comparison model without the LLM guesses by setting $\alpha=\gamma=0$ in the two objectives above (and reselecting $\lambda$ by LOOCV, etc.).

To estimate $\theta$, we use a residual-on-residual estimator. Define cross-fitted residuals for all $i$ by
\begin{equation*}
\hat r_i \;=\; Y_i - \widehat g^{(-k(i))}(W_i),\qquad
\hat v_i \;=\; D_i - \widehat m^{(-k(i))}(W_i),
\end{equation*}
where $k(i)$ is the fold containing observation $i$. The DML estimator of $\theta_0$ is the slope from regressing $\hat r$ on $\hat v$ without intercept:
\begin{equation*}
\hat\theta \;=\; \frac{\sum_{i=1}^n \hat v_i \hat r_i}{\sum_{i=1}^n \hat v_i^2}\,.
\end{equation*}
To characterize uncertainty, we report heteroskedasticity-robust standard errors as the square root of the sandwich estimator,
\begin{equation*}
\widehat{\mathrm{Var}}(\hat\theta)
\;=\;
\left(\sum_{i=1}^n \hat v_i^2\right)^{-2}
\sum_{i=1}^n \hat v_i^2 \,\hat\varepsilon_i^{\,2},
\qquad
\hat\varepsilon_i \;=\; \hat r_i - \hat\theta\,\hat v_i.
\end{equation*}

 %To estimate the effect of feedback score on price, we employ the partially linear model, taking $D$ to be the quantile of the feedback score and $Y$ to be the log of the auction price. To estimate $\E[Y|W]$, we use a LASSO regression of $Y$ on the embeddings as well as the LLM prediction $\tilde Y$, but leaving the coefficient on $\tilde Y$ unregularized. We performed an outer five-fold sample split with an inner leave-one-out cross-validation (LOOCV) for hyperparamter selection. That is, we partitioned the sample into folds $K_1,...,K_5$. On each fold $K_l$, we ran LOOCV to select our \texttt{glm} hyperparameter $\lambda_l$. Finally, we applied the chosen penalty fit to the model on $\hat\E_l[\cdot |W]$ on the training set $K_l^c$. We then made out-of-sample predictions $\hat \E_l[Y|W_i]$ on $i\in K_l$. For comparison purposes, we do the same with only the embeddings -- excluding the LLM guess entirely from the regression. Similarly, to estimate $\E[D|W]$, we use a LASSO regression of $D$ on the embeddings and $\tilde D$, leaving the coefficient on $\tilde D$ unregularized, with the penalty on each fold selected via LOOCV. Causal effect estimates are computed using residual-on-residual regression (using the out-of-fold predictions), with robust (sandwich) standard errors.

 \subsection{Results} We compare our approach to a comparison approach that excludes the LLM guess. Relative to the comparison approach, RMSE for the prediction of $Y$ (log price) decreases significantly; RMSE for the prediction of $D$ (feedback score quantile) does not change. The robust (sandwich) standard errors of the residual-on-residual regression shrinks, though neither estimate is statistically significant at conventional levels.  The correlations between ground truth and LLM prediction was $0.669$ and $0.500$, respectively. The improvement to prediction error for $Y$ is robust to transformation: indeed, without the log transformation on price, the comparative advantage of the LLM over the embeddings grows as it is better able to estimate the raw prices of very expensive jewelry. \\

\begin{table}[H]
    \centering
    \begin{tabular}{r|c c c c c }
         & RMSE ($\E[Y|W]$) & RMSE ($\E[D|W]$) & $\hat\theta$ & Robust SE \\
        Embeddings only & $0.8892$ & $0.1991$ & $-0.002429$ & $ 0.2484$  \\ 
        With LLM & $0.7357$ & $0.1985$ & $-0.052504$ & $0.1677$ 
    \end{tabular}
    \caption{Residual-on-residual regressions}
    \label{tab:placeholder}
\end{table}
Sample LLM outputs are presented in \cref{sampleoutputs}. An important observation is that the reasoning used to produce a guess of $\E[Y|W]$ reproduces standard economic reasoning by first estimating a melt value, and using that to calibrate predictions. The same cannot be said for $\E[D|W]$, where we are unable to attain similar efficiency gains.

% we believe the actual data-generating process (DGP) of $Y$ to be, whereas the same cannot be said for $D$. Specifically, a reasonable guess for the true DGP of individual bids $b_i$ is that the buyer uses their own chain-of-thought reasoning to estimate a melt value for the jewelry, 
% \begin{align*}
%     \log b_j=Q_{0}(W)+\underbrace{\xi_j+\varepsilon_j}_{\eta_j}
% \end{align*}
% where $Q_0$ is the melt value of the jewelry, and $\xi_j$ is the buyer's estimation error, and $\varepsilon_j$ is the idiosyncratic taste shock. Then, the DGP of $Y$ is 
% \begin{align*}
%     Y=Q_0(W)+\max_{j\leq n} \eta_j  && \E[Y|W]=Q_0(W)+\int \E[\max_{j\leq n} \eta_j]\ \mathbb{P}(dn)
% \end{align*}

\section{Discussion}

Here we interpret the empirical and methodological findings, explain why directly incorporating LLM-generated guesses of $\E[Y|W]$ and $\E[D|W]$ can tighten double/debiased ML estimators, and outline extensions.

\subsection{Why might this work?}

Generating causally relevant synthetic variables from LLMs trained on historical data (e.g., a direct guess of $\E[Y\mid W]$ or $\E[D\mid W]$) uses the model’s world knowledge as a prior to partially “unravel” nonlinear relationships that are hard to learn from limited $n$. This can outperform raw embeddings, which are not invertible and may discard task-critical detail; prompting also steers the model’s hidden representations toward the causal target without fine-tuning. Related work pursues a complementary ``alignment" strategy by tailoring architectures or embeddings for DML-style nuisance learning. \\

Furthermore, the reasoning capacities of modern LLM-powered AI tools may help in settings where outcomes or treatments are generated by human reasoning. In such cases, a low-dimensional summary of that reasoning can proxy the functions $\E[Y\mid W]$ and $\E[D\mid W]$. In auctions, for example, bidders plausibly reason toward a melt-value–anchored valuation; an LLM’s chain of thought performs a similar computation, yielding a monotone signal for the same latent construct. Under double/debiased machine learning (DML), improving these nuisance estimates tightens the final causal estimator—consistent with our application (lower RMSE for $E[Y\mid W]$ and smaller robust standard errors). \\

But even zero-shot predictions can help because they implicitly expose the model’s learned weights, reflecting priors about nonlinearities acquired from very large corpora. When the embedding dimension is comparable to $n$ and sparsity is limited, estimating those weights from scratch is perilous; importing the LLM’s implied weights provides a useful (if imperfect) prior that improves nuisance estimation.

\subsection{Extensions}

It is possible to construct other synthetic predictors to ease in nuisance function estimation. Reasoning-based LLMs generally produce these guesses through intermediate steps, reasoning about the latent variables (such as the weight of gold of the item) which may affect $Y$ or $D$. We propose extending this work to include direct guesses of intermediate latent variables as predictors. The choice of these variables can be guided by researcher knowledge or through querying the LLM itself. These may provide some robustness to distributional shift, in case the relationship between these latent variables and outcomes may be different in the researcher's sample than in the historical data with which the LLM was trained. \\

Another natural extension would be to include direct LLM predictions in other classes of model and estimator. For example, we could consider a less restrictive model than the partially linear model, or we could use an explicitly nonlinear estimator to replace the LASSO. We could also directly incorporate these predictions into an integrated supervised workflow \parencite{klaassen2024doublemldeep} so as to align the fine-tuning process toward residual nonlinearity.

\printbibliography

\section{Appendix}

\subsection{Prompt for Inferring Final Price}\label{prompt_price}
\noindent 
You are an e-commerce due-diligence analyst. Your job is to estimate the sale price in USD of an eBay
jewelry listing. \\

\noindent
Rules:\\
- Think step by step. \\
- After your reasoning, output the final numeric guess wrapped EXACTLY like: <final>100</final>\\
- Never include more than one <final> tag. \\
- Do not include dollar sign or anything other than the number. \\

\subsection{Prompt for Inferring Feedback Score}\label{prompt_fs}
\noindent 
    You are an e-commerce analyst specializing in fine-jewelry auctions on eBay. Your job is to estimate the seller’s feedback score, defined as:
  feedback\_score = (\# positive ratings) - (\# negative ratings)\\

\noindent Context: Feedback score correlates with customer satisfaction and—critically—with seller scale/experience.\\

\noindent ESTIMATION HEURISTICS (guidance, not hard rules):\\
- Very Low:     0–50       → one-off/private sellers; amateur photos; no returns; sparse specifics.\\
- Low:          50–500     → small casual sellers; mixed photo quality; some policy text.\\
- Medium:       500–5,000  → established storefront; consistent SKUs/watermarks; standard returns; many specifics.\\
- High:         5,000–50,000 → power sellers; professional studio images; branded store; fast/paid shipping options.\\
- Ultra:        50,000+    → top eBay stores; highly standardized listings; extensive policies; certification partners.\\

\noindent PRICE BAND (inferred from purity/weight/brand/gemstones/certification/condition/images):\\
- <\$100, \$100–\$500, \$500–\$2k, \$2k–\$10k, $>\$$10k\\
Use the band as a *soft* upward/downward adjustment: higher price bands tend to imply higher seller scale.\\

\noindent REASONING \& OUTPUT RULES\\
- Provide 2–5 terse evidence bullets naming only the strongest *observable* signals (include the inferred price band).\\
- Be specific (“studio lighting + branded watermark” instead of “good images”).\\
- If a signal is absent/unclear, say “no signal” rather than guessing.\\
- Then output ONLY the final integer guess for the feedback score, wrapped EXACTLY once:\\
  <final>1234</final>\\
- Never include more than one <final> tag. Do not add any other text after the <final>…</final> line.\\
- The final must be an integer (round your estimate).\\

\noindent FORMAT\\
1) Bullets (2–5)\\
2) <final>…</final>

\subsection{Sample Outputs}\label{sampleoutputs}
\noindent 
Sample output 1 (auction price guess):\\
\texttt{- Item: 14K yellow gold pendant with small pear/oval emerald center surrounded by tiny diamonds; listed weight 1.1 g (pre-owned, excellent).\\
- Gold value: 1.1 g × 14K $\approx$ 0.64 g pure gold → scrap $\approx$ \$40–45 (at current spot).  \\
- Gem value: small emerald and melee diamonds (no certification) — modest value, roughly \$30–\$90 combined depending on quality.  \\
- Market factors: unbranded, poor/standard photos, estate find listing style, seller has strong feedback. Comparable small 14K emerald/diamond cluster pendants on eBay typically sell in the \~\$80–\$150 range.  \\
- Conclusion: reasonable expected sale (used, unbranded, uncertified) near the midpoint of comps.\\
<final>110</final>}\\

\noindent 
Sample output 2 (feedback score guess):\\ 
\texttt{- Amateur smartphone photos on a textured surface, no watermark or studio lighting — individual seller signal.\\
- Short, confusing description (“I believe gold plated” vs 14k marking) and unbranded item — low professionalism.\\
- Basic item specifics filled and condition listed as ""Pre-owned - Excellent"" — minor signal of some eBay familiarity.\\
- No visible store/returns information or fast/shipping options in listing metadata — no signal for a larger store.\\
- Inferred price band: \$100–\$500 (small gold charm $\sim$2.5 g)\\
<final>18</final>}\\

\subsection{Summary statistics}

% Table created by stargazer v.5.2.3 by Marek Hlavac, Social Policy Institute. E-mail: marek.hlavac at gmail.com
% Date and time: Thu, Oct 02, 2025 - 8:11:19 PM
\begin{table}[!htbp] \centering 
  \caption{Summary statistics} 
  \label{summary} 
\begin{tabular}{@{\extracolsep{5pt}}lccccccc} 
\\[-1.8ex]\hline 
\hline \\[-1.8ex] 
Statistic & \multicolumn{1}{c}{Mean} & \multicolumn{1}{c}{St. Dev.} & \multicolumn{1}{c}{Min} & \multicolumn{1}{c}{Pctl(25)} & \multicolumn{1}{c}{Median} & \multicolumn{1}{c}{Pctl(75)} & \multicolumn{1}{c}{Max} \\ 
\hline \\[-1.8ex] 
price & 265.05 & 599.37 & 0.99 & 64.76 & 108.49 & 242.50 & 7,450.50 \\ 
log price & 4.81 & 1.15 & $-$0.01 & 4.17 & 4.69 & 5.49 & 8.92 \\ 
feedback score & 10,056.56 & 58,547.04 & 0 & 621 & 4,062 & 9,823 & 1,041,349 \\ 
\# bids & 11.64 & 11.96 & 1 & 2 & 8 & 16 & 58 \\ 
\# images & 7.64 & 5.12 & 1 & 4 & 6 & 9 & 24 \\ 
\# chars. in text & 7,058.18 & 29,356.99 & 686 & 1,229 & 1,724 & 7,578 & 519,166 \\ 
price guess & 290.82 & 535.58 & 12.99 & 110 & 150 & 250 & 5,500 \\ 
score guess & 2,087.80 & 8,723.62 & 12 & 300 & 1,200 & 2,100 & 120,000 \\ 
\hline \\[-1.8ex] 
\end{tabular} 
\end{table} 

\end{document}